\documentclass{article}

\usepackage{microtype}
\usepackage{graphicx}
\usepackage{booktabs}

\usepackage[accepted]{icml2025}
\makeatletter
\renewcommand{\ICML@appearing}{\textit{Preprint. Work in progress.}}
\makeatother

\usepackage{amsmath}
\usepackage{amssymb}
\usepackage{mathtools}
\usepackage{enumitem}
\usepackage{xcolor}
\usepackage{algorithm}
\usepackage{algorithmic}
\usepackage{multirow}
\usepackage{tikz}
\usepackage{pgfplots}
\pgfplotsset{compat=1.14}

\newcommand{\tracer}{\textsc{Tracer}}
\newcommand{\ta}{\text{TA}}
\newcommand{\cov}{\text{Cov}}

\icmltitlerunning{TRACER: Trace-Based Adaptive Cost-Efficient Routing for LLM Classification}

\begin{document}

\twocolumn[
\icmltitle{TRACER: Trace-Based Adaptive Cost-Efficient Routing\\for LLM Classification}

\icmlsetsymbol{equal}{*}

\begin{icmlauthorlist}
\icmlauthor{Adam Rida}{ind}
\end{icmlauthorlist}

\icmlaffiliation{ind}{Independent Researcher}
\icmlcorrespondingauthor{Adam Rida}{adamrida.ra@gmail.com}

\vskip 0.3in
]

\printAffiliationsAndNotice{}

\begin{abstract}
Every call to an LLM classification endpoint produces a labeled input--output pair already retained in production logs.
These pairs constitute a free, growing training set: a lightweight surrogate trained on them can absorb a significant portion of future traffic at near-zero marginal inference cost.
The open questions are \emph{when} the surrogate is reliable enough to deploy, \emph{what} it handles versus defers, and \emph{how} that boundary evolves as data accumulates.

We introduce \tracer{} (\textbf{Tr}ace-based \textbf{A}daptive \textbf{C}ost-\textbf{E}fficient \textbf{R}outing), an open-source system that trains ML surrogates on an LLM's own production traces and governs deployment through a \emph{parity gate}: the surrogate is activated only when its agreement with the LLM exceeds a user-specified threshold~$\alpha$.
To make the routing boundary transparent, \tracer{} generates interpretability artifacts describing which input regions the surrogate handles, where it plateaus, and why it defers.

On a 77-class intent benchmark with a Sonnet~4.6 teacher, \tracer{} achieves 83--100\% surrogate coverage depending on the quality target~$\alpha$; on a 150-class benchmark, the surrogate fully replaces the teacher.
On a natural language inference task, the parity gate correctly refuses deployment because the embedding representation cannot support reliable separation.
The system is available as open-source software.\footnote{Code and documentation: \texttt{github.com/adrida/tracer}}
\end{abstract}

\vspace{4pt}

\section{Introduction}
\label{sec:intro}

Production teams increasingly use LLMs as zero-shot classifiers to avoid the cost and delay of manual data labeling~\citep{chae2025,sun2023}.
The appeal is clear: a single prompt replaces months of annotation work.
But every LLM classification call is billed, and at scale the per-query cost adds up quickly.
Each call, however, also produces a side effect: a labeled input--output pair, already logged for billing and compliance, that can serve as training data.
We call each such pair a \emph{trace}.
In our experiments, a surrogate trained on Sonnet~4.6 traces handles 100\% of 150-class intent traffic (157 effective classes including teacher label noise) with sub-millisecond CPU inference.

The \emph{learning-to-defer} (L2D) paradigm~\citep{madras2018,mozannar2020} formalizes this idea: route easy inputs to a cheap surrogate, defer hard ones to the teacher.
But existing L2D and LLM-routing methods~\citep{ding2024,ong2025,chen2023frugalgpt} require labeled data collected \emph{before} deployment.
\tracer{} removes this requirement: when the router defers to the teacher, the teacher's response \emph{is} the label.
Although fewer deferrals mean less new training data, each remaining deferral provides signal from the surrogate's uncertain regions, creating a self-reinforcing \textbf{teacher-trace flywheel}.

This flywheel raises two deployment questions.
\textbf{Safety:} when is the surrogate reliable enough to go live?
\tracer{} answers with a \emph{parity gate} that activates the surrogate only when its agreement with the teacher on a held-out split exceeds a threshold~$\alpha$.
\textbf{Trust:} what does the surrogate handle versus defer?
At each refit, interpretability artifacts describe the routing boundary, drawing on slice discovery~\citep{chung2019}, counterfactual contrasts~\citep{wachter2017}, and differential model explanations~\citep{rida2023deltaxplainer}.

Prior XAI methods explain individual \emph{predictions}.
\tracer{}'s artifacts explain the \emph{routing partition}: a meta-level decision about which inputs the surrogate can reliably handle.

\paragraph{Contributions.}
\begin{enumerate}[leftmargin=*,itemsep=2pt,topsep=2pt]
\item An open-source, trace-driven routing system with parity-gated deployment and a continual learning flywheel that requires no manual labeling.
\item An alpha sweep across two intent benchmarks (77 and 157 classes) showing the coverage--quality tradeoff, with a negative result on NLI confirming the gate's safety properties.
\item A set of per-refit interpretability artifacts (slice summaries, boundary pairs) that describe the deferral partition rather than individual predictions.
\end{enumerate}

\section{Related Work}
\label{sec:related}

\paragraph{Learning to defer.}
\citet{madras2018} and \citet{mozannar2020} frame deferral as joint classifier--rejector optimization; \citet{mozannar2023} add consistent surrogate losses; \citet{narasimhan2022} study post-hoc estimators.
All assume ground-truth labels collected upfront, which is often the main bottleneck in production settings where labeling budgets are limited and label schemas evolve over time.
\tracer{} sidesteps this requirement entirely by acquiring supervision from teacher traces accumulated during production.

\paragraph{LLM routing and cascades.}
The cascade framework~\citep{dohan2022cascades} formalizes sequential model invocation.
FrugalGPT~\citep{chen2023frugalgpt} cascades LLMs by cost; Hybrid~LLM~\citep{ding2024} routes by predicted difficulty; RouteLLM~\citep{ong2025} uses preference data; AutoMix~\citep{madaan2024automix} uses self-verification.
These route \emph{between LLMs}; \tracer{} routes from an LLM \emph{to classical ML}, a strictly cheaper target.

\paragraph{XAI for routing boundaries.}
Slice Finder~\citep{chung2019} discovers underperforming data slices; counterfactual explanations~\citep{wachter2017} highlight minimal input changes that flip a decision; DeltaXplainer~\citep{rida2023deltaxplainer} explains differences between models across time via interpretable decision rules.
These methods target classifier predictions.
\tracer{} adapts them to the \emph{deferral partition}: explaining what the surrogate handles versus defers.

\paragraph{Summary.}
\tracer{} combines ideas from L2D, LLM routing, and XAI into a single system that (i)~requires no upfront labeled data, (ii)~routes to classical ML rather than a cheaper LLM, and (iii)~makes the routing boundary inspectable.
To the best of our knowledge, no prior system integrates all three.

\section{Method}
\label{sec:method}

\subsection{Problem Setting}

A production classifier $T\!:\!\mathcal{X}\!\to\!\mathcal{Y}$ (LLM API) generates a trace $(x, T(x))$ per call.
\tracer{} fits a surrogate $f$ and acceptor $a\!:\!\mathcal{X}\!\to\![0,1]$, producing a hybrid classifier $h$:
\begin{equation}
h(x)=\begin{cases}f(x)&\text{if } a(x)\geq\tau,\\T(x)&\text{otherwise},\end{cases}
\end{equation}
where $h$ routes each input to either the surrogate (handled) or the teacher (deferred to the LLM).
The objective is to maximize \emph{coverage}, $\cov\!=\!\Pr[a(x)\!\geq\!\tau]$, subject to \emph{teacher agreement} (TA), $\ta\!=\!\Pr[f(x)\!=\!T(x)\mid a(x)\!\geq\!\tau]\!\geq\!\alpha$.

\subsection{Routing Pipeline}

Surrogates operate on precomputed text embeddings.
At each refit, \tracer{} trains a pool of candidate classifiers (logistic regression, SGD linear classifiers, MLPs, decision trees, random forests, and extra-trees ensembles) and selects the best by teacher-label macro-F1 on a held-out validation split.

For each candidate surrogate, a binary \emph{acceptor} estimates whether the surrogate will agree with the teacher on a given input.
The acceptor is a logistic regression trained on four confidence features derived from the surrogate's output probability vector: the top-1 class probability, the top-2 probability, the margin between them, and the normalized entropy.
A natural alternative is to threshold the surrogate's \texttt{predict\_proba} output directly (i.e., defer when the maximum class probability is below some threshold).
We use a separate acceptor because it can combine multiple signals of uncertainty: a high top-1 probability with a small margin (two classes competing) is less reliable than the same top-1 probability with a large margin, and the acceptor learns this distinction.
In practice, the acceptor provides tighter TA guarantees than raw probability thresholding at the same coverage level (see Banking77 at $\alpha\!=\!0.95$ in Table~\ref{tab:main}, where \tracer{} achieves 0.959 TA vs.\ the baseline's 0.951).
The threshold~$\tau$ is calibrated on a held-out split by sweeping all unique acceptor scores and selecting the value that maximizes coverage while maintaining $\ta\!\geq\!\alpha$.

Two pipeline families compete at each refit:
\textbf{Global}, where a single surrogate handles all traffic without acceptor gating (eligible only when the surrogate's overall TA already exceeds~$\alpha$); and
\textbf{L2D} (learning-to-defer), where the surrogate is paired with the acceptor gate so that only high-confidence predictions are served by the surrogate.
The system selects whichever family achieves higher coverage while meeting the TA constraint.
(The codebase also implements a residual two-stage cascade; it was never selected as optimal in our experiments and is omitted here for clarity.)

\subsection{Parity Gate}

Before a candidate pipeline is promoted to production, it must pass a \emph{parity gate} evaluated on a held-out \emph{shadow split} never used for training or threshold calibration.
The candidate achieving the highest coverage while meeting $\ta\!\geq\!\alpha$ is promoted only if it also clears a 5\% coverage floor (preventing degenerate pipelines).

Two properties make this design robust.
First, it is \emph{conservative}: if no candidate clears the gate, the system falls back to full teacher reliance.
Second, it is \emph{temporal}: a pipeline that fails on day~$t$ may pass on a later day once the label buffer contains more traces.

\subsection{Continual Learning (The Flywheel)}

On day~1, every input is sent to the teacher, and all resulting traces enter the label buffer.
\tracer{} fits a first candidate pipeline and applies the parity gate.
On subsequent days, the active pipeline routes traffic: high-confidence inputs are handled by the surrogate, low-confidence inputs are deferred to the teacher.

Each call to \texttt{tracer.update()} merges the new traces and refits from scratch, ensuring that threshold calibration remains valid on the full accumulated dataset.
The flywheel requires no manual labeling: every deferred call produces a free training example, naturally biased toward the decision boundary where the surrogate needs the most signal.

\subsection{Interpretability Artifacts}
\label{sec:artifacts}

At each refit, \tracer{} generates five artifact types describing the \emph{deferral boundary}:

\begin{enumerate}[leftmargin=*,itemsep=3pt,topsep=2pt,label=(\Alph*)]
\item \textbf{Slice summaries.}
  For each teacher-assigned class label (and optionally for length bins), the system reports what fraction of traffic is handled by the surrogate versus deferred to the teacher, along with the per-slice TA.
  An operator can immediately see which categories the surrogate owns and which remain teacher-dependent without inspecting individual predictions.
  Grounded in automated slice discovery~\citep{chung2019}.

\item \textbf{Representative example cards.}
  For each routing group (handled and deferred) within each class, the system selects the input closest to the embedding centroid of that group.
  These examples give the operator a concrete mental model of what ``typical handled traffic'' and ``typical deferred traffic'' look like in practice.

\item \textbf{Contrastive boundary pairs.}
  Pairs of inputs that share the \emph{same} teacher label but receive \emph{opposite} routing decisions: one handled with a high acceptor score, one deferred with a low score.
  By comparing two inputs that differ only in routing outcome, the operator can identify what feature of the input (phrasing ambiguity, topic sensitivity, lexical overlap with other classes) causes the surrogate to defer.
  Grounded in counterfactual explanations~\citep{wachter2017}.

\item \textbf{Temporal deltas.}
  After a refit, per-label changes in handled rate compared to the previous routing state (available from the second refit onward).
  The operator can audit whether each refit is expanding coverage (positive delta) or regressing (negative delta), and which specific categories are moving.
  Inspired by DeltaXplainer's~\citep{rida2023deltaxplainer} approach to explaining how model behavior changes over time.

\item \textbf{Disagreement cards.}
  Cases where the surrogate and teacher disagree on held-out data, grouped by the surrogate's predicted class.
  For tasks that pass the parity gate, these highlight residual failure modes.
  For tasks that never pass the gate, they explain \emph{why} the surrogate fails (e.g., the surrogate defaults to one class, revealing a representation gap).
\end{enumerate}

\section{Experiments}
\label{sec:experiments}

\subsection{Setup}

We run \texttt{tracer.fit()} and \texttt{tracer.update()} from the shipped open-source package.
All inputs are embedded with BGE-large-en-v1.5~(1024 dimensions), precomputed offline.

\paragraph{Teacher.}
All tasks use \textbf{Claude Sonnet~4.6} as the teacher LLM.
Cached teacher predictions are used throughout (no live API calls during evaluation).

\paragraph{Tasks.}
\begin{itemize}[leftmargin=*,itemsep=1pt,topsep=2pt]
\item \textbf{Banking77}~\citep{casanueva2020}: 77-class intent classification.
  10{,}003 train traces, 3{,}080 test. Teacher accuracy: 78.7\% (train), 81.0\% (test).
\item \textbf{CLINC150}~\citep{larson2019}: 150 in-scope intent classes.
  18{,}000 train traces, 4{,}500 test. Teacher accuracy: 94.4\% (train), 93.1\% (test).
  The teacher generates 7 spurious labels not in the ground-truth label set, producing 157 effective classes in the trace data; the surrogate learns to reproduce these.
\item \textbf{MNLI}~\citep{williams2018}: 3-class natural language inference (entailment, neutral, contradiction).
  Used as a negative control with ground-truth labels as a stand-in teacher (a generous upper bound).
  We subsample 2{,}000, 5{,}000, and 10{,}000 training traces from the full 392K-example dataset; 9{,}815 test.
\end{itemize}

\paragraph{Alpha sweep.}
Rather than selecting a single $\alpha$ per task, we sweep $\alpha\!\in\!\{0.80, 0.85, 0.90, 0.95\}$ on each benchmark, mapping the coverage--quality tradeoff.

\paragraph{Protocol.}
Training data is divided into 5 equal daily batches.
Day~1: \texttt{tracer.fit()} on the first batch.
Days~2--5: \texttt{tracer.update()} with each subsequent batch.
Final evaluation on the held-out test set.

\paragraph{Baseline.}
\emph{Confidence-threshold deferral}: a logistic regression trained on \emph{all} traces at once (not incrementally), deferring inputs below a probability threshold swept to maximize coverage at each $\alpha$.
This baseline represents the best a simple deferral method can achieve with full hindsight; \tracer{}'s flywheel must reach this level starting from zero data.

\subsection{Main Results}
\label{sec:results}

\begin{table}[t]
\centering
\small
\caption{Alpha sweep results on held-out test sets. Coverage (Cov) and teacher agreement (TA) for \tracer{} and the confidence-threshold baseline. GT Acc is end-to-end accuracy against ground truth (surrogate on handled traffic, teacher on deferred). $^\dagger$Test TA below target $\alpha$ despite passing calibration gate (see \S\ref{sec:discussion}).}
\label{tab:main}
\begin{tabular}{@{}l c c c c c c@{}}
\toprule
& & \multicolumn{2}{c}{\textbf{\tracer{}}} & \multicolumn{2}{c}{\textbf{Baseline}} & \\
\cmidrule(lr){3-4} \cmidrule(lr){5-6}
\textbf{Task} & $\alpha$ & \textbf{Cov} & \textbf{TA} & \textbf{Cov} & \textbf{TA} & \textbf{GT} \\
\midrule
\multirow{4}{*}{Banking77}
 & .80 & 100\% & .894 & 100\% & .899 & .804 \\
 & .85 & 100\% & .894 & 100\% & .899 & .804 \\
 & .90 & 96.1\% & .912 & 99.8\% & .900 & .804 \\
 & .95 & 83.2\% & .959 & 87.8\% & .951 & .816 \\
\midrule
\multirow{4}{*}{CLINC150}
 & .80 & 100\% & .930 & 100\% & .942 & .924 \\
 & .85 & 100\% & .930 & 100\% & .942 & .924 \\
 & .90 & 100\% & .930 & 100\% & .942 & .924 \\
 & .95 & 100\% & .930$^\dagger$ & 98.4\% & .951 & .924 \\
\midrule
MNLI & .85--.95 & 0\% & -- & 0\% & -- & -- \\
\bottomrule
\end{tabular}
\end{table}

Table~\ref{tab:main} reveals three distinct regimes.

\paragraph{Regime 1: Full offload with quality headroom (CLINC150).}
The surrogate achieves 100\% coverage at all four alpha levels, including $\alpha\!=\!0.95$.
This means a logistic regression on BGE embeddings completely replaces Sonnet~4.6 on 150-class intent classification.
The parity gate selects the Global method (no acceptor needed) because the surrogate's calibration-set agreement (95.2\%) exceeds the threshold.
However, the test-set TA is 93.0\%, below the $\alpha\!=\!0.95$ target despite clearing it on the calibration split (95.2\%).
This cal--test gap is an important caveat: the parity gate's guarantee is only as strong as the representativeness of the calibration split.
In practice, practitioners should monitor test-time TA and re-calibrate periodically.
Ground-truth end-to-end accuracy is 92.4\%, comparable to the teacher-only baseline (93.1\%).
The confidence-threshold baseline also achieves higher TA on CLINC150 (0.942 vs.\ 0.930), reflecting the fact that the simpler method calibrates better on this particular task.

\paragraph{Regime 2: Coverage--quality tradeoff (Banking77).}
Banking77 is the more revealing benchmark.
At $\alpha\!\leq\!0.85$, the system achieves 100\% coverage (Global method).
At $\alpha\!=\!0.90$, the L2D pipeline activates: coverage drops to 96.1\% as the acceptor gate defers 3.9\% of traffic to maintain 91.2\% teacher agreement.
At $\alpha\!=\!0.95$, coverage drops further to 83.2\% with 95.9\% TA.
This is the alpha knob working as designed: the operator chooses the quality bar, and the system adjusts how much traffic it handles.

The confidence-threshold baseline, which trains on all 10{,}003 traces at once, achieves comparable or slightly higher coverage (87.8\% vs.\ 83.2\% at $\alpha\!=\!0.95$; 99.8\% vs.\ 96.1\% at $\alpha\!=\!0.90$).
This is expected: the baseline has the advantage of hindsight, seeing the complete dataset in a single batch.
\tracer{}'s flywheel reaches similar performance starting from 2{,}001 traces on Day~1, with the parity gate ensuring quality at every step.
The system's value is not that it beats a hindsight baseline on a static split, but that it (i)~begins offloading traffic from the first day of production, (ii)~gates each refit with a formal quality check, and (iii)~generates artifacts explaining the routing boundary.
At $\alpha\!=\!0.95$, \tracer{} achieves tighter TA than the baseline (0.959 vs.\ 0.951), reflecting the acceptor's ability to selectively defer uncertain inputs rather than relying solely on classifier confidence.

\paragraph{Regime 3: Correct refusal (MNLI).}
On 3-class NLI, the parity gate correctly refuses deployment at all alpha levels and all data scales tested (2{,}000, 5{,}000, and 10{,}000 ground-truth-labeled traces; $\alpha\!\in\!\{0.85,0.90,0.95\}$).
In all nine configurations, coverage is exactly 0\%: no candidate pipeline clears the gate.
This is a generous test (ground-truth labels rather than noisy teacher labels), yet the surrogate still cannot learn the entailment/neutral/contradiction boundary from BGE embeddings alone.
NLI requires compositional reasoning over premise--hypothesis pairs that a frozen sentence embedding cannot capture.
The gate's refusal is the correct behavior: it prevents a bad surrogate from going live.
Extending \tracer{} to tasks like NLI, where frozen embeddings are insufficient, is a natural direction for future work; fine-tuning the encoder on accumulated traces could unlock new task families while the parity gate continues to serve as the safety check.

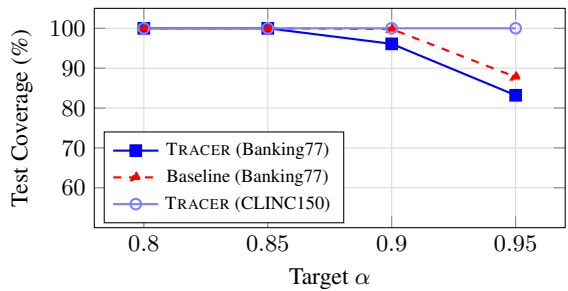
\begin{figure}[t]
\centering
\small
\begin{tikzpicture}
\begin{axis}[
    width=0.95\columnwidth,
    height=4.5cm,
    xlabel={Target $\alpha$},
    ylabel={Test Coverage (\%)},
    xmin=0.78, xmax=0.97,
    ymin=50, ymax=105,
    xtick={0.80,0.85,0.90,0.95},
    ytick={60,70,80,90,100},
    legend style={at={(0.02,0.02)},anchor=south west,font=\scriptsize},
    grid=major,
    grid style={gray!30},
]
\addplot[mark=square*,thick,blue] coordinates {(0.80,100) (0.85,100) (0.90,96.1) (0.95,83.2)};
\addlegendentry{\tracer{} (Banking77)}
\addplot[mark=triangle*,thick,red,dashed] coordinates {(0.80,100) (0.85,100) (0.90,99.8) (0.95,87.8)};
\addlegendentry{Baseline (Banking77)}
\addplot[mark=o,thick,blue!50] coordinates {(0.80,100) (0.85,100) (0.90,100) (0.95,100)};
\addlegendentry{\tracer{} (CLINC150)}
\end{axis}
\end{tikzpicture}
\caption{Coverage vs.\ target $\alpha$ on held-out test sets. Banking77 shows a clear tradeoff: raising $\alpha$ from 0.85 to 0.95 reduces coverage from 100\% to 83\%. CLINC150 achieves 100\% at all targets.}
\label{fig:alpha}
\end{figure}

\subsection{Flywheel Dynamics}

\begin{table}[t]
\centering
\small
\caption{Banking77 flywheel at $\alpha\!=\!0.95$ (the strictest setting). Coverage grows from 73\% to 83\% as traces accumulate, with TA at or above the target.}
\label{tab:flywheel}
\begin{tabular}{@{}c r c c@{}}
\toprule
\textbf{Day} & \textbf{Traces} & \textbf{Cov (cal)} & \textbf{TA (cal)} \\
\midrule
1 & 2{,}001  & 73.4\% & .953 \\
2 & 4{,}001  & 78.2\% & .951 \\
3 & 6{,}002  & 74.5\% & .951 \\
4 & 8{,}002  & 83.4\% & .950 \\
5 & 10{,}003 & 82.6\% & .951 \\
\midrule
\multicolumn{2}{l}{\emph{Test set}} & 83.2\% & .959 \\
\bottomrule
\end{tabular}
\end{table}

Table~\ref{tab:flywheel} shows the flywheel at its most demanding setting ($\alpha\!=\!0.95$).
With 2{,}001 traces on Day~1, the surrogate handles 73.4\% of 77-class intent traffic while maintaining TA above 0.95.
Coverage grows to 83.4\% by Day~4 as the surrogate sees more of the input distribution.
The small dip on Day~3 (74.5\%) reflects the refit-from-scratch recalibration; coverage recovers once more data arrives.

At lower alpha targets, the flywheel converges faster.
At $\alpha\!=\!0.80$, Banking77 reaches 100\% coverage on Day~1 from 2{,}001 traces alone.
CLINC150 at $\alpha\!=\!0.95$ starts at 57.4\% coverage on Day~1 (3{,}600 traces) and reaches 100\% by Day~2 (7{,}200 traces), with calibration TA dropping from 0.973 to 0.951 as the system shifts from the conservative L2D pipeline to the full-coverage Global pipeline.

\paragraph{Cost projection.}
At Sonnet~4.6 pricing (\$3/M input, \$15/M output tokens), Banking77 costs approximately \$2.60 per 1{,}000 teacher calls (estimating ${\sim}$800 input tokens for the prompt including 77 intent names, ${\sim}$15 output tokens).
A system routing 10K queries/day spends \$26/day (\$9{,}500/year).
With \tracer{} at 83.2\% coverage ($\alpha\!=\!0.95$), this drops to \$4.40/day, an 83\% cost reduction saving \$7{,}900/year.
At $\alpha\!=\!0.80$ (100\% coverage), the cost drops to zero after trace collection, a 100\% reduction.
For CLINC150, the surrogate fully replaces the teacher at all alpha levels, eliminating ongoing LLM costs entirely.

\subsection{Artifact Inspection}
\label{sec:xai-results}

We inspect the interpretability artifacts to check whether they produce sensible output across the three regimes.
This is a qualitative inspection; a rigorous evaluation of artifact utility would require a user study with practitioners, which we leave to future work.
We include the inspection here to illustrate the artifact design and to show that the outputs are non-trivial.

\paragraph{CLINC150 (full offload).}
Slice summaries show 100\% handled rate across all 157 classes with uniform TA.
No boundary pairs are generated (none exist at 100\% coverage), which is the correct artifact behavior: the output reflects the routing state, not a fixed template.
The Global method selection is itself an interpretable signal: the surrogate is good enough everywhere that gating adds no value.

\paragraph{Banking77 at $\alpha\!=\!0.95$ (partial offload).}
At 83.2\% coverage, slice summaries reveal per-class variation in handled rates.
Handled rates range from 78.2\% (\texttt{card\_payment\_not\_recognised}) to 96.4\% (\texttt{transaction\_charged\_twice}), showing that the surrogate is weaker on intents with overlapping surface forms.
Length has a modest effect: short and medium inputs are handled at 90.7--91.0\%, while long inputs drop to 88.3\%.

Boundary pairs show semantically meaningful contrasts.
Table~\ref{tab:boundary} lists three representative pairs from the Banking77 qualitative report.
In each case, the handled input is direct and unambiguous, while the deferred input uses vaguer phrasing that could plausibly belong to a neighboring intent class.
The acceptor scores reflect this: handled inputs score 0.94--0.96, while deferred inputs score 0.00.

\begin{table}[t]
\centering
\scriptsize
\caption{Contrastive boundary pairs from Banking77 at $\alpha\!=\!0.95$. Each pair shares the same teacher label but opposite routing decisions. Full output in Appendix~\ref{app:artifacts}.}
\label{tab:boundary}
\begin{tabular}{@{}p{2.6cm} p{2.6cm} p{1.6cm}@{}}
\toprule
\textbf{Handled} (score) & \textbf{Deferred} (score) & \textbf{Label} \\
\midrule
``I returned an item but don't see it on my account?'' (0.95) & ``Why hasn't my return cleared my account?'' (0.00) & refund\_not\_ showing\_up \\
\addlinespace
``What is the procedure for activating this card?'' (0.96) & ``How do I get started when I get my card?'' (0.00) & activate\_ my\_card \\
\addlinespace
``Are there ATM fees?'' (0.94) & ``Do any of your machines provide cash from my home country?'' (0.00) & atm\_support \\
\bottomrule
\end{tabular}
\end{table}

\paragraph{MNLI (correct refusal).}
No qualitative report is generated because the parity gate never fires at any of the nine tested configurations.
The absence of artifacts is itself informative: it confirms there is no viable routing boundary.
Unlike Banking77, where the surrogate learns intent-specific clusters, MNLI's 3-class structure (entailment, neutral, contradiction) cannot be separated from sentence-pair embeddings because the class depends on the \emph{logical relation} between premise and hypothesis, not on the topic or phrasing of either sentence alone.
The artifacts direct the practitioner toward a richer representation.

Table~\ref{tab:xai} summarizes the artifact signatures across regimes.

\begin{table}[t]
\centering
\small
\caption{Artifact signatures across routing regimes.}
\label{tab:xai}
\begin{tabular}{@{}p{1.3cm} p{2.8cm} p{2.5cm}@{}}
\toprule
\textbf{Regime} & \textbf{Artifact signature} & \textbf{Operator action} \\
\midrule
Full offload & All slices 100\%. No boundary pairs. Global selected. & Ship. Monitor for drift. \\
\addlinespace
Partial & Uneven slices. Boundary pairs show ambiguity-based deferral. & Review deferred classes. Collect more traces. \\
\addlinespace
Refused & No artifacts. Gate never fires. & Task needs richer encoder or is unsuitable for surrogates. \\
\bottomrule
\end{tabular}
\end{table}

\section{Discussion}
\label{sec:discussion}

\paragraph{When \tracer{} works.}
The results show that intent classification, even at 77 or 157 classes, is well-served by frozen embeddings plus classical ML.
This is because intent tasks have relatively clean class boundaries in embedding space.
Recent surveys confirm that LLM-based zero-shot classification is increasingly adopted in production as a way to bypass manual annotation~\citep{chae2025,sun2023}.
The practical implication is that many of these systems are paying per-call LLM prices for tasks that classical ML can handle, and \tracer{} provides a principled, safe way to offload that traffic.

\paragraph{Calibration--test gap.}
On CLINC150, the parity gate passes at $\alpha\!=\!0.95$ based on calibration TA (0.952), but test TA is 0.930.
This gap means the gate's guarantee does not fully transfer to unseen data.
The most likely cause is that the calibration and test splits have slightly different difficulty distributions.
A practical mitigation is to set $\alpha$ conservatively above the desired test-time TA, or to use conformal calibration~\citep{angelopoulos2021} for distribution-free coverage guarantees.
We leave conformal integration to future work.

\paragraph{When it does not.}
MNLI's failure demonstrates a clear boundary.
Tasks requiring compositional reasoning (entailment vs.\ contradiction) produce embeddings where class boundaries are not linearly separable, regardless of data volume.
The parity gate is essential here: without it, a practitioner might deploy a bad surrogate.
A natural extension is fine-tuning the encoder on accumulated traces, which could extend the viable task range while the parity gate continues to serve as the safety check.

\paragraph{Surrogate imitates teacher, not ground truth.}
The surrogate replicates the teacher's behavior, including its errors.
On Banking77, Sonnet~4.6 achieves only 78.7\% train / 81.0\% test ground-truth accuracy across 77 fine-grained intents (e.g., distinguishing \texttt{card\_not\_working} from \texttt{contactless\_not\_working}).
This reflects the genuine difficulty of zero-shot 77-way classification, not a prompt deficiency.
At $\alpha\!=\!0.80$ (100\% coverage), end-to-end GT accuracy is 80.4\%, close to the teacher's 81.0\% on the test set.
At $\alpha\!=\!0.95$ (83.2\% coverage), GT accuracy rises to 81.6\% because the hardest inputs are deferred back to the teacher, matching the teacher-only baseline.
\tracer{} does not claim to improve on the teacher; it claims to \emph{match} it at a fraction of the cost.
An interesting corollary is that the deferred inputs, which the surrogate cannot confidently classify, constitute a natural hard-sample set.
These are precisely the inputs where human annotation effort would be most valuable, suggesting that \tracer{}'s deferral log could double as a cost-efficient active labeling strategy: instead of labeling random samples, practitioners could focus annotation budgets on the deferred partition to build a ground-truth dataset where it matters most.

\paragraph{Limitations and future work.}
The current evaluation points to concrete directions for future development.
\emph{Broader task coverage:} both positive benchmarks are intent classification tasks with well-separated embedding clusters. Extending to tasks with noisier labels, longer inputs, multi-label structure, or compositional reasoning (as MNLI suggests) will likely require encoder fine-tuning on accumulated traces, which the parity gate can govern safely.
\emph{Stronger baselines:} the confidence-threshold comparator is the simplest possible deferral method. Planned comparisons include learned routing methods (e.g., RouteLLM~\citep{ong2025}) and distillation approaches~\citep{hsieh2023distilling}.
\emph{Embedding diversity:} all experiments use BGE-large-en-v1.5; the coverage--quality tradeoff may shift with different encoders, and a systematic encoder comparison is planned.
\emph{Artifact evaluation:} the XAI artifacts are inspected qualitatively here; a user study with practitioners would provide stronger evidence of their utility in real deployment workflows.
\emph{Incremental updates:} the system currently refits from scratch at each update; incremental fitting would reduce compute cost for large trace buffers.
\emph{Conformal guarantees:} replacing the empirical calibration threshold with conformal prediction~\citep{angelopoulos2021} would provide distribution-free coverage guarantees, addressing the calibration--test gap observed on CLINC150.

\section{Conclusion}

LLMs are best understood not as classifiers but as \emph{labeling engines}: their most valuable output is the training signal each call produces for downstream surrogates.
\tracer{} operationalizes this insight.
On Banking77 (77 classes), the alpha knob controls the coverage--quality tradeoff from 100\% coverage at $\alpha\!=\!0.80$ down to 83.2\% at $\alpha\!=\!0.95$.
On CLINC150 (157 classes), the surrogate fully replaces the teacher at all quality targets.
On MNLI, the parity gate correctly refuses deployment, demonstrating safety by design.

The system is available as open-source at \texttt{github.com/adrida/tracer}.

\bibliography{tracer}
\bibliographystyle{icml2025}

\appendix
\section{Teacher Prompts}
\label{app:prompts}

All teacher calls use Claude Sonnet~4.6 with structured output (JSON schema).

\paragraph{Banking77.}
\begin{quote}\small
\texttt{Classify the following banking customer query into exactly one of these intent categories: [77 intents listed]. Return JSON: \{"label": "<intent>"\}.}
\end{quote}

\paragraph{CLINC150.}
\begin{quote}\small
\texttt{Classify the following user utterance into exactly one of these intent categories: [150 in-scope intents listed]. Return JSON: \{"label": "<intent>"\}.}
\end{quote}
\noindent The teacher occasionally generates labels outside the provided 150-class set (7 spurious labels such as \texttt{hotel\_launchpad}, \texttt{pause\_music}), producing 157 effective classes in the trace data.

\section{Full Artifact Output}
\label{app:artifacts}

\paragraph{Banking77 boundary pairs ($\alpha\!=\!0.95$).}
All five contrastive pairs from the qualitative report:

\begin{small}
\begin{enumerate}[leftmargin=*,itemsep=4pt]
\item \textbf{Refund\_not\_showing\_up.}
  Handled (0.95): ``I returned an item but don't see it on my account?''
  Deferred (0.00): ``Why hasn't my return cleared my account?''

\item \textbf{activate\_my\_card.}
  Handled (0.96): ``What is the procedure for activating this card?''
  Deferred (0.00): ``How do I get started when I get my card?''

\item \textbf{apple\_pay\_or\_google\_pay.}
  Handled (0.96): ``Can I use my apple watch to top-up?''
  Deferred (0.00): ``I got my American Express in Apple Pay, why is top up not working on my device?''

\item \textbf{atm\_support.}
  Handled (0.94): ``Are there ATM fees?''
  Deferred (0.00): ``Do any of your machines provide cash from my home country? I don't have any money on me\ldots''

\item \textbf{automatic\_top\_up.}
  Handled (0.96): ``i need help finding the auto top up option.''
  Deferred (0.00): ``How can I Use thereto-top option?''
\end{enumerate}
\end{small}

\paragraph{Banking77 slice summaries ($\alpha\!=\!0.95$).}
Selected label slices ordered by handled rate:

\begin{small}
\begin{tabular}{@{}l r r r@{}}
\toprule
\textbf{Label} & \textbf{Handled} & \textbf{TA} & $n$ \\
\midrule
\texttt{card\_payment\_not\_recognised} & 78.2\% & .961 & 197 \\
\texttt{top\_up\_failed} & 80.4\% & .955 & 194 \\
\texttt{declined\_card\_payment} & 83.2\% & .970 & 197 \\
\midrule
\texttt{activate\_my\_card} & 93.4\% & .959 & 181 \\
\texttt{cash\_withdrawal\_charge} & 95.0\% & .995 & 202 \\
\texttt{transaction\_charged\_twice} & 96.4\% & .994 & 169 \\
\bottomrule
\end{tabular}
\end{small}

\end{document}